%% file: main.tex
\documentclass[conference,twocolumn]{IEEEtran}

\usepackage{cite} % Tidies up citation numbers.
\usepackage{url} % Provides better formatting of URLs.
\usepackage[utf8]{inputenc} % Allows Turkish characters.
\usepackage{booktabs} % Allows the use of \toprule, \midrule and \bottomrule in tables for horizontal lines
\usepackage{graphicx}

\usepackage{subcaption}
\usepackage{booktabs}
\usepackage{threeparttable}
\usepackage{amsmath, amssymb}

\usepackage{soul,color}
\definecolor{Red}{cmyk}{0,1,1,0}
\definecolor{Green}{cmyk}{1,0,1,0}
\definecolor{Cyan}{cmyk}{1,0,0,0}
\definecolor{Purple}{cmyk}{0.45,0.86,0,0}
\definecolor{Rosolic}{cmyk}{0.00,1.00,0.50,0}
\definecolor{Blue}{cmyk}{1.00,1.00,0.00,0}
\definecolor{Orange}{cmyk}{0,0.52,0.80,0}
\definecolor{Black}{cmyk}{1,0,0,1}

\newcommand{\shortcite}[1]{\cite{#1}}
\newcommand{\sysName}[0]{DeepPortraitDrawing}

\begin{document}

%%%%%%%%% TITLE
\title{\sysName:\\ Generating Human Body Images from Freehand Sketches
%FashionDrawing\hbc{DeepHumanDrawing? Fashion can mean the clothes only? You may define a latex command for this short name}: Generating Human Images from Hand-Drawn Sketches
}

\author{Xian Wu,
        Chen Wang,
        Hongbo Fu,
        Ariel Shamir,
        Song-Hai Zhang and 
    %   Song-Hai Zhang\IEEEauthorrefmark{1}% <-this % stops a space
    Shi-Min Hu\IEEEauthorrefmark{1}
    \IEEEcompsocitemizethanks{ \IEEEcompsocthanksitem Xian Wu, Chen Wang and
      Shi-Min Hu(corresponding author, e-mail:
shimin@tsinghua.edu.cn) are with the Department of Computer Science and Technology, Tsinghua University, Beijing, China.
\IEEEcompsocthanksitem Yi-Xiao Li is with the Academy of Arts \& Design, Tsinghua University, Beijing, China.}}

\maketitle

\input{abstract}

\input{introduction}
\input{related_work}
\input{method}
\input{experiments}
\input{conclusion}

\bibliographystyle{IEEEtran}
\bibliography{IEEEabrv,egbib}

\end{document}

%% file: abstract.tex
% !TEX root = ..\cvpr.tex

\begin{abstract}
   Researchers have explored various ways to generate realistic images from freehand sketches, e.g., for objects and human faces. %many ways to generate various images from input sketches, such as faces and objects. 
   However, how to generate realistic human body images from sketches is still a challenging problem. It is, first because of the sensitivity to human shapes, second because of the complexity of human images caused by body shape and pose changes, and third because of the domain gap between realistic images and freehand sketches.
   In this work, we present \sysName, a deep generative framework for converting roughly drawn sketches to  realistic human body images. To encode complicated body shapes under various poses, we take a local-to-global approach. Locally, we employ semantic part auto-encoders to construct part-level shape spaces, which are useful for refining the geometry of an input pre-segmented hand-drawn sketch. Globally, we employ a cascaded spatial transformer network to refine the structure of body parts by adjusting their spatial locations and relative  proportions. 
 Finally, we use a global synthesis network for the sketch-to-image translation task, and a face refinement network to enhance facial details. Extensive experiments have shown that given roughly sketched human portraits, our method produces more realistic images %with more reasonable body structures 
 than the state-of-the-art sketch-to-image synthesis techniques.
\end{abstract}

%% file: introduction.tex
% !TEX root = ..\cvpr.tex

\section{Introduction}

\begin{figure}
\centering
\includegraphics[width=3.2in]{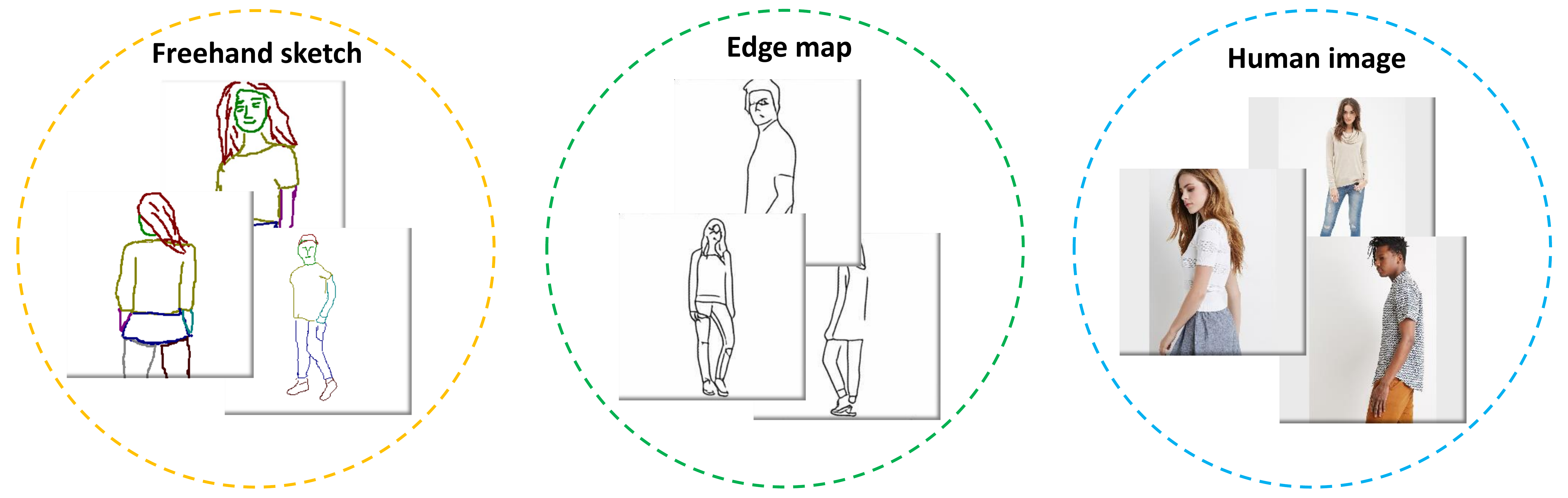}
\caption{There are huge gaps between %the 
freehand sketches with human images and the extracted edge maps. The freehand sketches, especially by those with few drawing skills, might not describe the local geometry or global structure of a human body accurately. }
\label{fig:teaser}
\end{figure}

Creating realistic human images benefits various applications, such as fashion design, movie special effects, and educational training. Generating human images from freehand sketches can be more effective since even non-professional users are familiar with such a pen-and-paper paradigm. % an effective way to achieve this goal and is also convenient for non-professional users. 
Sketches can not only represent the global structure of a
human body but also depict the local appearance details of the body as well as garments. 
%, \hb{including } which can meet users' all-around requirements.

Deep generative models, such as generative adversarial networks (GANs) \cite{goodfellow2014generative} and variational auto-encoders (VAEs) \cite{kingma2013auto}, have recently made a breakthrough for image generation tasks. Based on these generative models, many methods \cite{isola2017image,sangkloy2017scribbler,lu2018image,chen2018sketchygan} have been proposed to generate desired images from input sketches by solving a general image-to-image translation problem. Some other methods have focused on generating specific types of images, including human faces \cite{li2020deepfacepencil,chen2020deepfacedrawing} and foreground objects \cite{ghosh2019interactive}. Such methods can better handle freehand sketches by incorporating the relevant domain knowledge.
%which are even applicable for freehand sketches. 

Compared to many other types of images, human body images have more complicated intrinsic structures and larger shape and pose variations, 
making the sketch-based synthesis task difficult for the following reasons. First, existing human portrait image datasets \cite{liu2016deepfashion} only cover a small subset of all possible human images under various changing conditions of pose, shape, viewpoint, and garment. Since the existing sketch-to-image translation techniques often use pairs of images and their corresponding edge maps for training, they may fail to generate desired results when a test sketch is under very different conditions.
% different from any of the training edge maps
Second, hand-drawn sketches, especially those created by users with little drawing skills, can hardly describe accurate body geometry and structure, and look very different from edge maps extracted from the training images (Figure \ref{fig:teaser}).

In this work, we present \textit{\sysName}, a novel deep generative approach for generating realistic human images from coarse, rough freehand sketches (Figure~\ref{fig:pipeline}). Instead of trying to increase the generalization ability of sketch-to-image algorithms, our key idea is to project an input test sketch to %a 
part-level shape spaces constructed based on image-based training data. This can assist to bridge the gap between the training and test data, and also  the gap between freehand sketches and realistic images. This idea makes sense for our task since roughly drawn sketches do not provide hard constraints for geometric interpretation. By properly recombining part-level information in different training images we are able to cover a significant portion of all possible human images.

%On the one hand, the generated images have to represent exact semantics for the human body and each semantic component needs to have \hb{plausible geometry and} coherent local textures. On the other hand, the generated images have to \hb{reflect the structure of a human body}, %build reasonable topology for the human body,
%meaning that the body proportion and the joint connections should obey the physiological property of the human beings. However, hand-drawn sketches, \hb{especially for those from users with little drawing skill}, can hardly describe accurate \hb{body geometry and structure, making general-purpose sketch-to-image synthesis approaches difficult for synthesizing realistic human images from freehand sketches.} 
%\hbc{The above discussion has not touched the challenges caused by garments?}

%semantic information and credible body structures, therefore general sketch-to-image synthesis approaches cannot produce plausible results for human images.

%In order to tackle this challenge, we present a novel \hb{deep generative approach for generating realistic human images from coarse, rough freehand sketches \hbc{(Figure xxx)}. 

To this end, we take a local-to-global approach to encode complicated body shapes under various poses. For each semantic body part, we employ an auto-encoder to define a part-level latent shape space by training on part-level edge maps extracted from images. Our system takes as input a semantically segmented sketch, whose individual body parts are projected onto the constructed part-level shape spaces. This results in a geometrically refined sketch map and a corresponding parsing map (i.e., labeled regions). Next, 
we employ a cascaded spatial transformer network to structurally refine the sketch and parsing maps by adjusting the locations and relative proportions of individual body parts. Finally, we use a global synthesis network to produce a realistic human image from the transformed maps, and use a face refinement network to improve the local details of facial landmarks.

%so that we can separately address the geometry inference of individual body parts and the structure inference involving all body parts.
 
%sketch-based generation method for the human image, to meet the novice users' drawing requirements in practice. Firstly, we segment the input sketch according to the part label and feed the part sketches into the corresponding auto-encoders. The encoded latent vectors are projected into the underlying manifold, and then decoded to the sketch maps and the parsing maps which conform to the potential geometric shape. Secondly, we transform the decoded maps to reconstruct the global topology for the human body, according to the guidance from the predicted pose keypoints. Finally, the global synthesis network produces the human image from the transformed maps, and the face refinement network furthermore improves the local details of the facial landmarks.

Extensive experiments demonstrate the effectiveness and practicability of our method. We are able to satisfy novice users' need  for creating visually pleasing human images from hand-drawn sketches. In our self-collected dataset of freehand sketches, our method produces visually more pleasing results with more realistic local details, compared to the previous sketch-based image generation techniques (Figure \ref{fig:comparsion}). % and more reasonable body structures. 
The main contributions of our paper can be summarized 
as follows: 

\begin{itemize}
    \item We are the first to consider the problem of synthesizing realistic human images from roughly drawn sketches; 
    \item We present a local-to-global deep generative solution to geometrically and structurally refine an input %sketch 
    sketched human before image synthesis. 
    \item We collect a hand-drawn sketch dataset of human images (containing 308 segmented sketches), which can facilitate future research. 
    % \hbc{Maybe briefly discuss how big the dataset is.}
%    bridge the domain gap between freehand sketches and realistic images. 
\end{itemize}
%\hbc{I rewrote the contributions here since the original three contributions seem not significant individually.}

% into three folds:
% \begin{itemize}
% \item we extract the semantic structures from the input sketch by the part-based auto-encoders to guide the image synthesis;
% \item we reconstruct the human body topology through the affine transformations, according to the connection relationships of the pose keypoints;
% \item we employ the global synthesis network and the face refinement network for the realistic generation, making it possible for non-artists to draw satisfying human images.
% \end{itemize}

%% file: related_work.tex
% !TEX root = ..\cvpr.tex

\section{Related Work}
% \asc{I changed the grammar of this whole section to present tense instead of past tense (e.g. `retrieved' to `retrieve')}
\subsection{Sketch-to-image generation}

Generating desired images from hand-drawn sketches is a difficult task, since sketches often exhibit different levels of abstraction. To address this domain gap, traditional methods take a \textit{retrieval-composition} approach, essentially considering sketches as soft constraints. For example, a pioneering work by Chen et al. \shortcite{chen2009sketch2photo}
% \shortcite{chen2009sketch2photo}
% \hbc{Here you need to use `shortcite'. Please fix the same issues in the paper.} 
first retrieves images from the Internet using input sketches with text descriptions, and fuses the retrieved foreground and background images into desired pictures. A similar idea is used in PhotoSketcher \cite{eitz2011photosketcher}.
%with similar geometric shapes to the sketches, 
%and then 
%utilized the retrieved foreground and background images to composite the desired pictures. PhotoSketcher \cite{eitz2011photosketcher} retrieved images by sketches in real-time, so that users could interactively edit and refine the generated images. 
PoseShop \cite{chen2013poseshop} constructs image scenes with human figures but requires users to provide 2D poses for retrieval. Since such retrieval-based approaches directly reuse portions of existing images for re-composition, their performance is highly dependent on the scale of image datasets, as well as the composition quality.

By using deep learning models,  (e.g., conditional GANs \cite{mirza2014conditional}),  recent sketch-based image synthesis works adopt a \textit{reconstruction}-based approach. Some works \cite{isola2017image,zhu2017unpaired,wang2018high} aim at general-purpose image-to-image translation and can handle sketches as one of the possible input types. % input. 
Other works focus on using sketches as the condition for GANs. For example,
%Above traditional methods applied image retrieval to composite the results, nevertheless, deep learning models were able to generate images from sketches directly. Image-to-image translation methods \cite{isola2017image,zhu2017unpaired,wang2018high} leveraged conditional GANs \cite{mirza2014conditional} to produce realistic images from input sketches. 
Scribbler \cite{sangkloy2017scribbler} can control textures in generated images by grayscale sketches and colorful strokes. Contextual-GAN \cite{lu2018image} updates latent vectors for input sketches through back propagation and produces images by a pre-trained model. SketchyGAN \cite{chen2018sketchygan} and iSketchNFill \cite{ghosh2019interactive} are able to generate multi-class images for diverse sketches by introducing gated conditions. Gao et al. \shortcite{gao2020sketchycoco} propose an approach to produce scene images from sketches, by generating each foreground object instance and the background individually.
Recently, Ho et al. \shortcite{ho2020sketch} propose a coarse-to-fine generation framework and incorporate human poses to synthesize human body images. 
While impressive results were presented in the above works, these techniques do not generalize well to rough or low-quality sketches, which have very different characteristics compared to image edge-maps used for training the generative models. Additionally, since sketches are largely used as hard constraints in these techniques, the synthesized images would inherit geometric distortions if they exist in the input sketches (Figure \ref{fig:comparsion}).

Our approach has been inspired by the recent work DeepFaceDrawing \cite{chen2020deepfacedrawing}, which takes %took 
a \textit{projection-reconstruction} approach for synthesizing realistic human face images from sketches. The key idea of DeepFaceDrawing is to refine the input sketches before synthesizing the final image. This refinement is achieved by projecting the input sketches to component-level spaces spanned by edge maps of realistic faces.
DeepFaceDrawing achieves impressive results even for rough or incomplete sketches but is %were 
limited to the synthesis of frontal faces. We extend their approach to synthesizing human body images under various poses and viewpoints. Our extension explicitly uses the semantic information in the whole pipeline, and contributes a spatial transformation module, essentially leading to a \textit{projection-transformation-reconstruction} pipeline.

%DeepFaceDrawing  separated the sketch into five facial components and projected each of them into the underlying manifold. DeepFaceDrawing \cite{chen2020deepfacedrawing} achieved impressive results for freehand sketches even drawn by novice users but limited to frontal faces. However, these methods are not applicable for human images as they cannot address the complicated body structures in various poses and shapes. \wx{, 
%which is limited to extracted edge maps but not adaptive to freehand sketches.}

\subsection{Label-to-image generation}

There are many semantic synthesis approaches generating images from segmentation label maps. For example, Pix2pix \cite{isola2017image} is a general image-to-image translation framework based on a U-Net \cite{ronneberger2015u} generator and a conditional discriminator. Chen and Koltun \shortcite{chen2017photographic} present a cascaded refinement network and use multi-layer perceptual losses to achieve photographic images from segmentation maps. Pix2pixHD \cite{wang2018high} employs multi-scale generators and discriminators, and incorporates a feature matching loss to build a high-resolution image-to-image translation framework. GauGAN \cite{park2019semantic} introduces the SPADE layer to control image styles directly by semantic segmentation. Zhu et al. \shortcite{zhu2020semantically} present a semantically multi-modal synthesis model to generate images with diverse styles for each semantic label. LGGAN \cite{tang2020local} combines local class-specific sub-generators and a global image-level generator for semantic scene generation. DAGAN \cite{tang2020dual} present two novel attention modules to capture spatial-wise and channel-wise attention individually.
% Different from the above parametric models\hbc{you consider the above approaches as parametric models? If yes, make it clearer}, Qi et al. \cite{qi2018semi} propose a semi-parametric image synthesis method. 
Different from the above \textit{reconstruction}-based approaches, Qi et al. \shortcite{qi2018semi} introduce a \textit{retrieval-reconstruction} image synthesis method.
They retrieve image segments from a dataset using segmentation maps as query and employ a global refinement network to produce globally consistent %realistic 
results. % in a global consistence.
Although segmentation labels can be used to generate plausible images, they are less expressive than sketches in describing local details and geometric textures of user-desired images. (e.g., collars and sleeves in Figure~\ref{fig:comparsion}) % \hbc{Can you refer to a specific example in the paper? Your current sketches mainly describe the boundary of each segment. It's better to have some results with interior strokes describing local details}

\subsection{Human body image generation}

Human-body image synthesis is challenging, because of human sensitivity to human shapes. There is a need to make the global body structure reasonable and to produce realistic local textures.
Most researchers have focused on the human pose transfer task \cite{ma2017pose,ma2018disentangled}, which synthesizes the same person from a source image in target poses. To achieve this, some methods utilize component masks \cite{balakrishnan2018synthesizing,siarohin2018deformable}, human parsing \cite{dong2018soft,han2019clothflow}, or correspondence flows \cite{li2019dense,Liu_2019_ICCV,ren2020deep} to transform %, transforming 
local source features into target areas, thus preserving %therefore preserved 
the appearance of the same person in target poses. Other methods \cite{lassner2017generative,neverova2018dense} employ dense pose \cite{alp2018densepose} or statistical human models like %introduced
SMPL \cite{loper2015smpl} %model
to provide the human body structure as a prior. Several methods \cite{neverova2018dense,Sarkar2020,Liu2019Neural} construct a surface texture map from a
source human body image, and then render the texture map on a target human image. Recently, HumanGAN \cite{sarkar2021humangan} proposes novel part-based encoding and warping modules for generating diverse human images with high quality. These pose transfer techniques focus on preserving texture details from source images, while our method focuses on generating body textures and garments according to hand-drawn sketches.

Besides pose, other approaches synthesize human images by modifying other properties.
%there are \hb{several} approaches \hb{for} synthesizing human images based on other properties. 
For example, FashionGAN \cite{Zhu_2017_ICCV} encodes the shape, appearance, and text, %the appearance, and the text, respectively,
allowing to edit garment textures of human images through text descriptions. Many researchers have attempted to
address the virtual try-on problem \cite{han2018viton,wang2018toward}, i.e., dressing
a source person with given clothes through %a 
proper geometric transformations.
%paid attention to the virtual try-on problem~\cite{han2018viton,wang2018toward} , \hb{i.e.,} to dress \hb{a} %the source person with given clothes through a geometric transformation. 
Ak et al. \shortcite{ak2019attribute} and Men et al. \shortcite{men2020controllable} use %the
attribute vectors to represent appearance information and then control the clothes and textures of human images via such attribute vectors. % for the human image. 
Dong et al. \shortcite{dong2020fashion} leverage a parsing map as guidance and introduce an attention normalization layer to edit human images by sketches and colors. These methods are able to change certain properties for a source human image, but they cannot generate a brand-new human image from scratch. % to satisfy users with a variety of demands.

%% file: method.tex
% !TEX root = ..\cvpr.tex

\begin{figure*}
\centering
\includegraphics[width=1.0\textwidth]{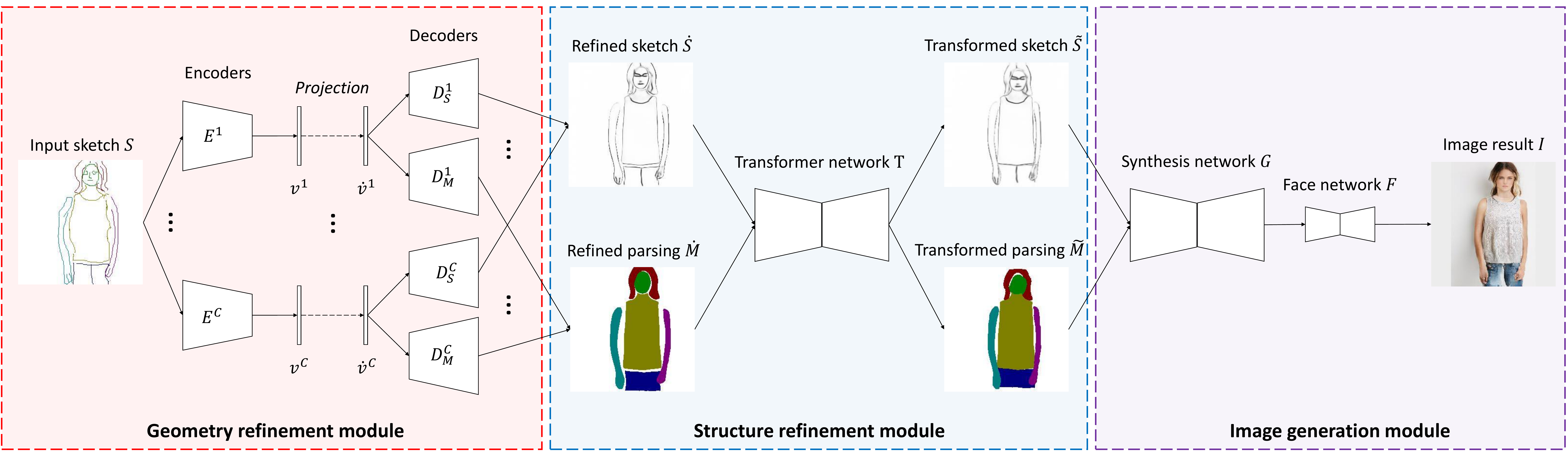}
\caption{Pipeline of the proposed \textit{projection-transformation-reconstruction} approach to generate human body images from freehand sketches. Firstly, individual body parts of an input sketch are projected onto the underlying part-level manifolds %the input sketch is projected into the underlying manifold 
and decoded into a geometrically refined sketch map and a parsing map, based on an auto-encoder architecture. Secondly, the individual parts of the refined sketch map and the parsing map are %affine 
transformed  via
a cascaded spatial transformer network, to refine the global structure of the human body. Thirdly, the transformed maps are fed into the global synthesis network to generate a new %the 
human image and then the face refinement network to enhance the facial details.}
\label{fig:pipeline}
\end{figure*}

\section{Method}

% \wxc{Now $S$ denotes the input sketch and $\dot S$ denotes the geometry-refined sketch. ($S_0$ and $S$ are used respectively before.)}

% \asc{I think that $\dot S$ and $\dot M$ are still difficult to discern from $S$ and $M$, I would use $\tilde S$ and $\tilde M$ or $\hat S$ and $\hat M$}

We propose a \textit{projection-transformation-reconstruction} approach for generating realistic human body images from freehand sketches. As illustrated in Figure \ref{fig:pipeline}, it is achieved through three modules operated in sequence: a geometry refinement module, a structure refinement module, and an image generation module. 
The geometry refinement module takes a semantically segmented sketch as input and refines the geometry of its individual body parts by retrieving and interpolating the exemplar body parts in the latent spaces of the learned part-level auto-encoders. This module results in a refined sketch map and a corresponding parsing map. The structure refinement module spatially transforms the sketch and parsing maps to better connect and shape individual parts, and refine the relative proportions of body parts. Finally, the image generation module translates the transformed maps into a realistic human body image.

\subsection{Geometry refinement module} % Sketch conversion module}

This module aims to refine an input freehand sketch by using human portrait images to train several part-level networks. This has two advantages.  First, locally pushing the input sketch towards the training edge maps, and second reducing the geometric errors in the input sketch. This assists the image generation module in generating more realistic images.

Due to the complexity of human images, it is very unlikely to find in our training dataset an image that is globally similar to an input sketch (Figure~\ref{fig:comparsion}). On the other hand, it is much easier to retrieve similar body parts and learn a component-level shape space for each body part. We thus follow the idea in DeepFaceDrawing \cite{chen2020deepfacedrawing} to perform manifold projection at the component level. 

DeepFaceDrawing has focused
%focuses 
on the synthesis of frontal faces and relies on a shadow interface to guide users to sketch face components that are well aligned with the training examples. This alignment is critical for synthesizing realistic faces with DeepFaceDrawing. 
In contrast, we aim to handle portrait images under various poses and viewpoints. Hence, we cannot use a single layout template for body components. Instead, we propose to use the semantic segmentation information through the entire pipeline, since semantic labels provide a natural way to establish corresponding body parts in different images.  

Let $S$ denote a test sketch or a training edge map. We assume that $S$ has been semantically segmented into $C=8$ parts, including hair, face, top-clothes, bottom-clothes, left and right arms, left and right legs. We denote the part sketches as $\{S^c\}_{c=1,...,C}$. Each body part $S^c$ is cropped by a corresponding bounding box ($S^c$ will be a white image if part-$c$ is absent from $S$). We use an auto-encoder architecture to extract a feature vector for each body part to facilitate the subsequent manifold projection task, as illustrated in Figure \ref{fig:pipeline}.

% As illustrated in Figure \ref{fig:pipeline}, each part sketch $S^c$ is fed into a corresponding encoder $E^c$ to obtain a latent vector $v^c$ that represents the geometric and semantic features of $S^c$. $v^c$ is then concurrently fed into two decoders, a sketch decoder $D_S^c$ and a mask decoder $D_M^c$. The former $D_S^c$ decodes the latent vector $v^c$ into a part sketch $S^c$\hbc{maybe add a tilde to indicate this is a reconstructed version of $S^c$}, while the latter decodes $v^c$ %the mask decoder $D_M^c$ decode it into a part mask $M^c$ indicating the part region. In the training stage, the $\{v^c_i\}$ collected from a set of training images can be considered as the samples that build the underlying component-level manifold for part $c$.

In the testing stage, given a semantically segmented sketch denoted as $\{S^c\}_{c=1,...,C}$, we project its body parts to the underlying %component
part-level manifolds for geometric refinement. We adopt the Locally Linear Embedding (LLE) algorithm \cite{roweis2000nonlinear} to perform manifold projection without explicitly constructing each part-level %component-level 
manifold. Specifically, each part sketch $S^c$ is first encoded into a latent vector $v^c$ by a corresponding encoder $E^c$. Based on the local linear assumption, we use a retrieve-and-interpolate approach. In more detail, we first retrieve $K$ nearest neighbors $\{v^c_k\}_{k=1,...,K}$ for $v^c$ in the latent space $\{v^c_i\}$ for part $c$ using the Euclidean distance. $\{v^c_i\}$ collected from a set of training images can be considered as the samples that build the underlying part-level manifold for part $c$. We then interpolate the retrieved neighbors to approximate $v^c$ by minimizing the mean squared error as follows:
%
% In the testing stage, given a semantically segmented sketch denoted as $\{S^c_0\}_{c=1,...,C}$, we project its body parts to the underlying component-level manifolds for geometric refinement. We adopt the Locally Linear Embedding (LLE) algorithm \cite{roweis2000nonlinear} to perform manifold projection without explicitly constructing each component-level manifold. Specifically, each part $S^c_0$ is first encoded into a latent vector $v^c_0$. Based on the local linear assumption, we use a retrieve-and-interpolate approach. In more detail, we first retrieve $K$ nearest neighbors $\{v^c_k\}_{k=1,...,K}$ for $v^c_0$ in the latent space for part $c$ using Euclidean distance, and then interpolate the retrieved neighbors to approximate $v^c_0$ by minimizing the mean squared error as follows:
%
\begin{equation}
    \min \|v^c - \sum_{k=1}^K w^c_k \cdot v^c_k\|_2^2, \quad s.t. \sum_{k=1}^K w^c_k=1,
\end{equation}
where $K=10$ in our experiments and $w^c_k$ is the unknown weight of the $k$-th vector candidate. For each body part, $\{w^c_k\}$ %w^c_k$ 
can be found independently by solving a constrained least-squares problem. After the weights $\{w^c_k\}$ are found, we can calculate the projected vector $\dot v^c$ by linear interpolation:
\begin{equation}
    \dot v^c = \sum_{k=1}^K w^c_k \cdot v^c_k.
\end{equation}

Next, the sketch decoder $D_S^c$ and the mask decoder $D_M^c$ for part $c$ process the projected vector $\dot v^c$, %respectively, 
resulting in a refined %projected
part sketch $\dot S^c$ and a part mask $\dot M^c$, respectively. Finally, all projected part sketches $\{\dot S^c\}$ and masks $\{\dot M^c\}$ are combined together to recover the global body shape, resulting in a geometry-refined sketch map $\dot S$ and a human parsing map $\dot M$.
% \hbc{$S$ has been used earlier. Consider using another symbol to avoid confusion. In addition, you may consider adding symbols into Figure 1. }  

% \wxc{I will rewrite all the network architecture details in Sec 4.2}

% \hb{\textbf{Network details.}}
% The part encoder $E^c$ contains five downsampling convolutional layers, with each downsampling layer followed by a residual block. A fully-connected layer is appended in the end to encode the features into the latent vector $v^c$ with $512$ dimensions. Similarly, the part decoders $D_S^c$ and $D_M^c$ contain five upsampling convolutional layers and five residual blocks in total. The final convolutional layer reconstructs the part sketch $S^c$ or the part mask $M^c$, accordingly.

% \hb{\textbf{Training.} }
In the training stage, 
we first train the encoder $E^c$ and the sketch decoder $D_S^c$ to avoid %, avoiding 
the distraction from the mask branch. Since $E^c$ and $D_S^c$ need to reconstruct the input $S^c$ with consistent shapes and fine details, we employ the $L_2$ distance as the reconstruction loss to train them. Then, we fix the weights of the parameters in $E^c$ and train the mask decoder $D_M^c$. We use the cross-entropy loss for this training since it is a binary segmentation task. %where $M^c$ is a binary segmentation mask. 

\subsection{Structure refinement module}

The geometry refinement module focuses only on the refinement of the geometry of individual body parts in a sketch. However, relative positions and proportions between body parts in a hand-drawn sketch might not be accurate. We thus employ the structure refinement module to refine the relative positions and proportions of body parts to get a globally more consistent body image. 

\begin{figure}
\centering
\includegraphics[width=0.5\textwidth]{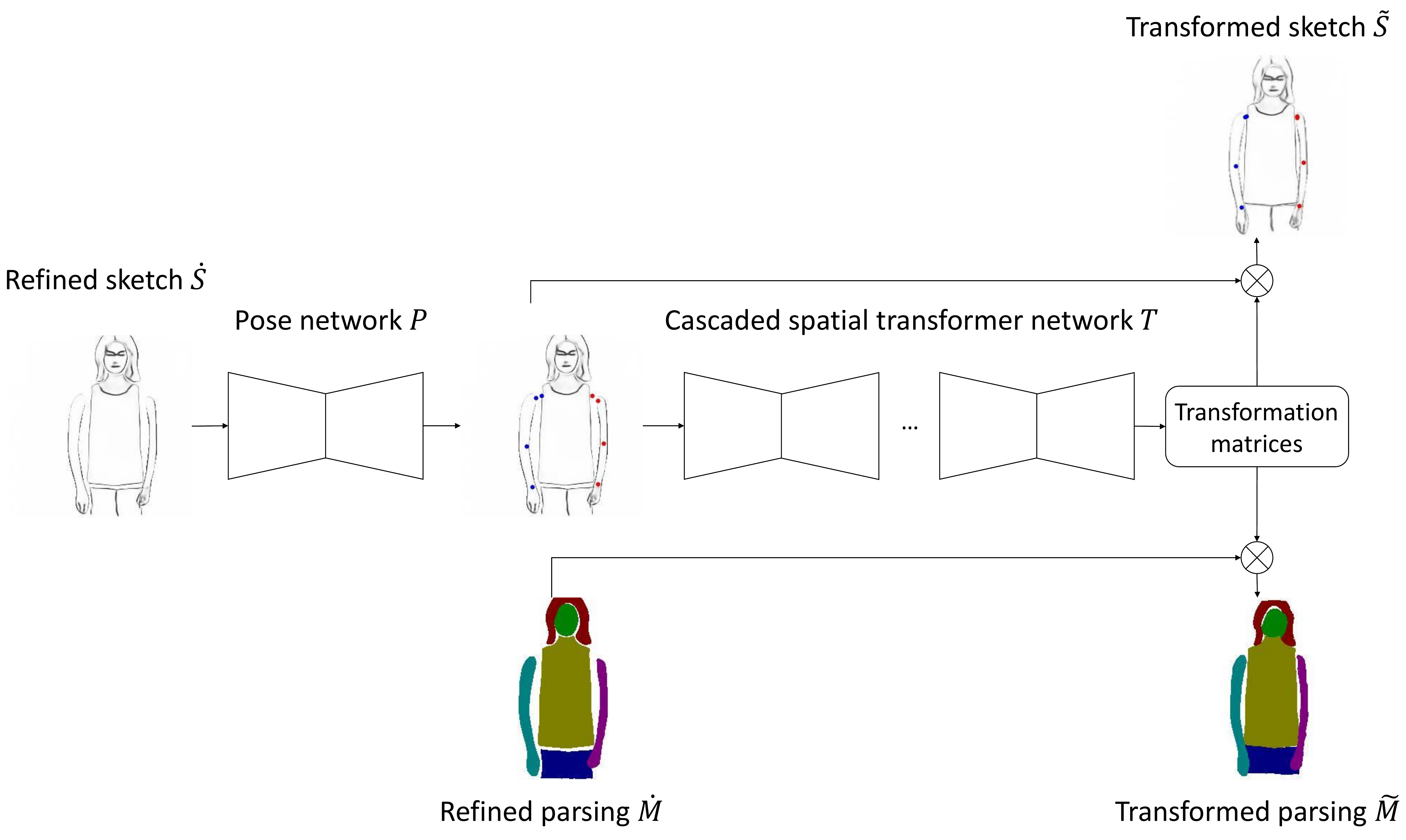}
\caption{Illustration of the structure refinement module. The keypoints of individual body parts (e.g., the arms and shoulders) are better connected and their relative length is globally more consistent after this step.}
\label{fig:structure}
\end{figure}

To refine the body structure, we use the pose keypoints (see Figure~\ref{fig:structure}), which provide a simple and effective way to represent a human body structure. According to the physiological characteristics of human beings, the positions of pose keypoints should obey two rules. First, a
joint of a body part should connect to the same joint of its neighboring body part. Second, the relative length of different body parts should be %as
{globally} consistent. Therefore, we aim to transform the keypoints of different body parts and make them conform to these rules.

% \hbc{Discuss the main idea first. Learn the distributions of the pose keypoints from the training images to refine the corresponding keypoints in an input sketch?}

As illustrated in Figure \ref{fig:structure}, we first utilize a pose estimation network $P$ to predict heatmaps $H^c$ %$\{H^c\}$ 
for the position of each keypoint from each refined part sketch map ${\dot S^c}$. % $\{\dot S^c\}$. 
Note that we need to predict the same joint repeatedly for neighboring body parts. Then, we leverage all the part heatmaps $\{H^c\}$ as guidance to recover the global structure of the sketched human body. The %However, the 
different body parts should preserve proper relative lengths, and connect with each other based on the inherent relationships among them. To achieve this, we apply affine transformations to the body parts predicted by a spatial transformer network \cite{jaderberg2015spatial} $T$, so that the part heatmaps $\{H^c\}$ are transformed to %a 
reasonable locations $\{\tilde H^c\}$  learned from real human poses. We apply the same predicted affine transformations to the refined part sketch maps %sketch map 
$\{\dot S^c\}$ and the part mask maps $\{\dot M^c\}$, resulting in $\{\tilde S^c\}$ and $\{\tilde M^c\}$, respectively. 

\begin{figure*}
\centering
\includegraphics[width=0.95\textwidth]{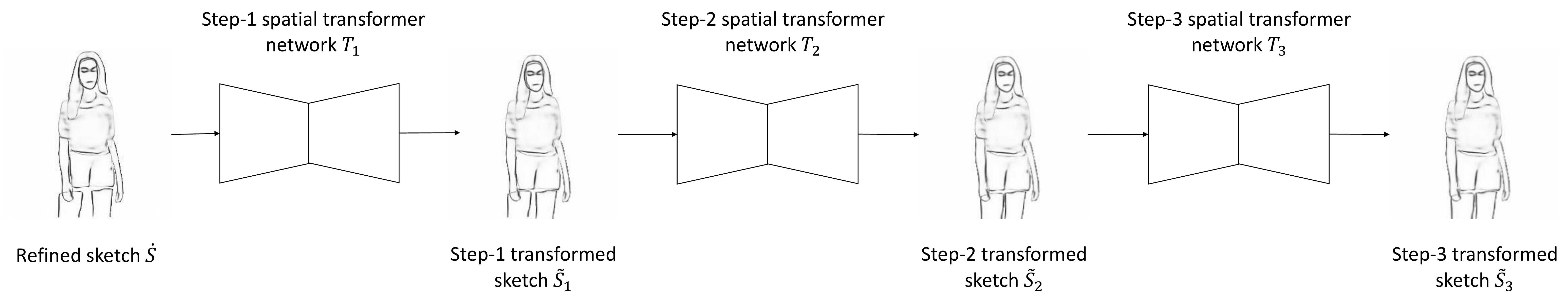}
\caption{In our experiments, a geometrically refined sketch map $\dot S$ is transformed iteratively for three steps to get a structurally refined sketch map.}
\label{fig:transformation}
\end{figure*}

Since neighboring body parts may influence each other, it is very difficult to recover the entire human structure in one step transformation. Therefore, we use a cascaded refinement strategy, employing a multi-step spatial transformer network to update the results iteratively. To leverage the global information, we combine all the part sketch maps as $\dot S$ and all the part heatmaps as $H$, and then feed $\dot S$ and $H$ to the spatial transformer network. The transformed sketch map $\tilde S$ and heatmaps $\tilde H$ in the $j$-th step are the input to the transformer network in the $(j+1)$-th step. In our experiments, we used a three-step refinement, as illustrated in Figure~\ref{fig:transformation}.

% \hbc{show the intermediate results in this iterative process? If you need more time, place them in the supplementary materials. This is important, since we emphasized the cascaded nature at multiple places in the paper}.

% and apply Equation \ref{eq:transform} to train all the transformer networks. 

%\hbc{In the abstract, it is said we use a cascaded spatial transformer. The names should be consistent throughout the paper.}

To train the pose estimation network $P$ and the cascaded spatial transformer network $T$, we need to simulate the inconsistencies of the global structure we may find at the test time. We apply random affine transformations to all part edge maps $\{S^c\}$ and part heatmaps $\{H^c\}$ in the training set, except for a selected
reference part. We select the top-clothes part (i.e., the upper body) as the reference part and keep it unchanged in our experiments. The pose network $P$ needs to predict all part heatmaps $\{\hat{H}^c\}$ 
% \hbc{If you follow AS's suggestion to update the symbols, make sure all the symbols are updated consistently.} %$\hat H$ 
from each randomly transformed edge map $\hat S$. We adopt the stacked hourglass architecture \cite{newell2016stacked} for $P$ and use the mean squared error to train it.

The goal of the cascaded spatial transformer network $T$ is to refine the size and location of each body part. Therefore, the predicted pose heatmaps $\{\hat H^c\}$ should be transformed so that they are as close to the ground-truth $\{H^c\}$ as possible. Similarly, we require the randomly transformed part edge maps $\{\hat S^c\}$ to be close to the ground-truth part edge maps $\{S^c\}$. %close to the ground-truth $\{\hat H^c\}$ after the transformation, as well as the sketches $\{S^c\}$.
%Besides that, 
We have found that extremely large transformations may lead to training instability. We thus append a regularization term to penalize transformation matrices that are too large.
% \hbc{Not very clear. Is it because the original location and size are roughly correct? So that's why you do not like to deviate too much from the original position and size?}
The spatial transformer network $T_{j+1}$ in the $(j+1)$-th step is fed with the transformed edge map $\hat S_j$ and the combined heatmaps $\hat H_j$ in the $j$-th step. Its %so the 
initial input is $\hat S_0$ and $\hat H_0$. The loss function of $T$ can be formulated as:
\begin{equation}
\begin{aligned}
    \mathcal{L}(T)=\sum_{j=0}^2 \sum_{c=1}^C &\lambda_H\|\mathcal{F}(T_{j+1}^c(\hat S_j, \hat H_j), \hat H_j^c) - H^c\|^2_2 
    \\ + &\lambda_S\|\mathcal{F}(T_{j+1}^c(\hat S_j, \hat H_j), \hat S_j^c) - S^c\|^2_2 
    \\ + &\lambda_L\|T_{j+1}^c(\hat S_j, \hat H_j) - \mathbb{1}\|^2_2,
\end{aligned}
\label{eq:transform}
\end{equation}
where $\mathcal{F}$ represents an affine transformation operation and $\mathbb{1}$ denotes the identity matrix. $T_{j+1}^c(\hat S_j, \hat H_j)$ denotes the predicted transformation matrix for the $c$-th body part in the $(j+1)$-th step. We set $\lambda_H=100$ and $\lambda_S=\lambda_L=1$ in our  experiment to balance the three terms.

% \wxc{I will rewrite all the network architecture details in Sec 4.2}
% The transformer network $T$ consists of five downsampling convolutional layers, five residual blocks, and last two fully-connected layers to predict the affine transformation matrices.

\subsection{Image generation module}

\begin{figure}
\centering
\includegraphics[width=0.48\textwidth]{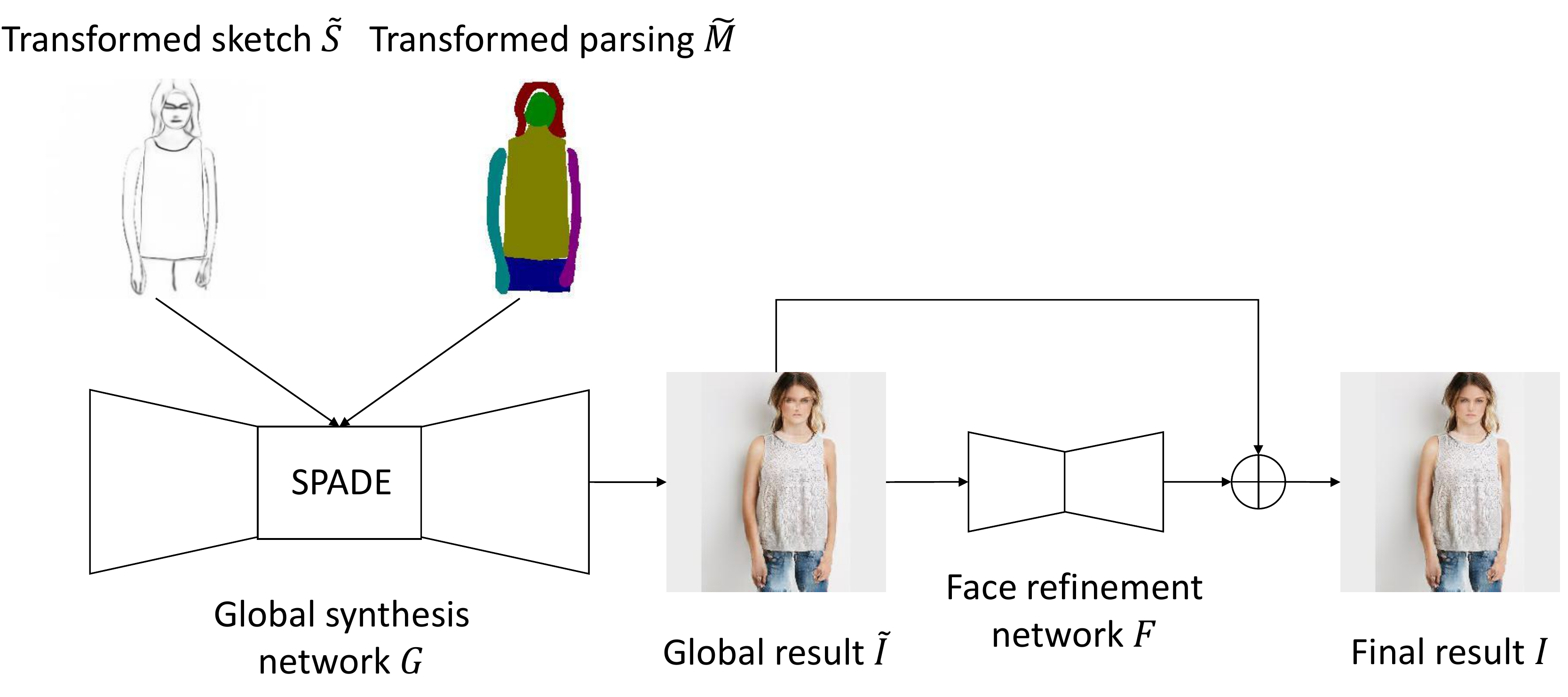}
\caption{Illustration of the image generation module. The transformed sketch and parsing maps are fed into the SPADE layer of the global synthesis network to produce a human image %global 
result. Then the face refinement network enhances the facial details for the final result.}
\label{fig:generation}
\end{figure}

Finally, we need to generate a desired human image $I$ from the transformed sketch map $\tilde S$ and the transformed parsing map $\tilde M$ after the structure refinement %topology reconstruction
module, as illustrated in Figure \ref{fig:generation}. We adopt GauGAN \cite{park2019semantic} as our basic architecture for the global synthesis network $G$, since it has achieved impressive results for the label-to-image translation task. The SPADE layer in GauGAN \cite{park2019semantic} takes the %segmentation 
parsing map $\tilde M$ as input by
default. To prevent losing the information in the sketch map $\tilde S$, we concatenate it to the parsing map $\tilde M$ and feed them together into the SPADE layer. This way, the parsing map $\tilde M$ controls the image style in each semantic region, while the sketch map $\tilde S$ provides the geometric features for local details.

The global synthesis network $G$ is able to generate an acceptable result $\tilde I$ globally. However, the human visual system is more sensitive to the quality of synthesized faces. Since hand-drawn human body sketches might not %rarely 
describe facial landmarks clearly, $G$ may fail to %not 
produce rich %realistic 
details for the face area. Inspired by Chan et al. \shortcite{chan2019dance}, we utilize a face refinement network $F$ to enhance the facial details in the human image $\tilde I$. We crop a square patch from $\tilde I$ according to the face label in $\tilde M$. The square patch and the face mask are then fed into the face refinement network $F$ to produce a %the 
residual image for the face area. The final result $I$ is the sum of $\tilde I$ and the residual image. To train $F$ to achieve a realistic human face, we adopt both an adversarial loss and a perceptual loss, similar to Chan et al. \shortcite{chan2019dance}.
% \hbc{are the losses here the same as those in \cite{chan2019dance}? Also how about the network architecture? If they are the same, explicitly say it to avoid any confusion.}

To train the global synthesis network $G$, we could simply take the edge maps $\{S_i\}$ 
% \hbc{why to make it boldface? The notations need to be cleaned up. In addition, it is not super clear what is the difference between the ground-truth sketch maps and the converted maps mentioned below. Can you have an illustration figure? Using the converted maps seems more straightforward according to the pipeline in Figure 1?} 
and the parsing maps $\{M_i\}$ in the training set as input. However, we have found that the synthesis network $G$ trained this way cannot address freehand sketches well. Although the geometry refinement %sketch conversion
module can refine the geometric shape of an input sketch $S$, the resulting sketch $\dot S$ still differs from edge maps found in the training set. The main reason is that edge maps extracted from natural human images
% human sketches \hbc{you mean the edge maps?} 
contain many texture details, and these can violate the local linear assumption \cite{roweis2000nonlinear} used in the step of manifold projection. Instead, to simulate the input at the test time, we take the projected version of each edge map in the training set as the input to train $G$. We retrieve $K$ nearest neighbors in the underlying manifold for each edge map $S_i$. % $\{S_i\}$. 
Then, the edge maps $\{\dot S_i\}$ and the parsing maps $\{\dot M_i\}$ decoded by the projected vectors are fed into $G$. Similar to GauGAN \cite{park2019semantic}, we adopt the adversarial loss, the perceptual loss, and the feature matching loss \cite{wang2018high} together to train $G$.

% \wxc{I will rewrite all the network architecture details in Sec 4.2}

%% file: experiments.tex
% !TEX root = ..\cvpr.tex

\section{Experiments}

To get the paired data for training, we construct a large-scale sketch dataset of human images from DeepFashion \cite{liu2016deepfashion}, as described in Sec \ref{sec:data}. Sec \ref{sec:implementation} introduces the architecture of our proposed networks and the implementation details of model training. We conduct comparison experiments with several sketch-to-image techniques in Sec \ref{sec:comparison} to show the superiority of our method for generating human images from hand-drawn sketches. The ablation study in Sec \ref{sec:ablation} evaluates the contribution of individual components in our method. Sec \ref{sec:multi-modal} shows that our method is able to produce multi-style human images from the same input sketches. 

\subsection{Data preparation}
\label{sec:data}

\begin{figure}[ht]
  \centering
    \begin{subfigure}{.115\textwidth}
    \includegraphics[width=0.8in]{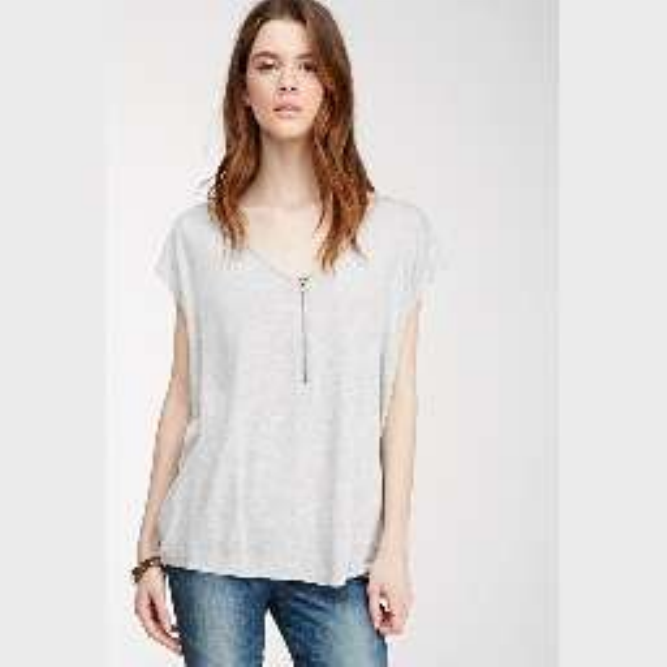}
    \caption{}
    \end{subfigure}
    \begin{subfigure}{.115\textwidth}
    \includegraphics[width=0.8in]{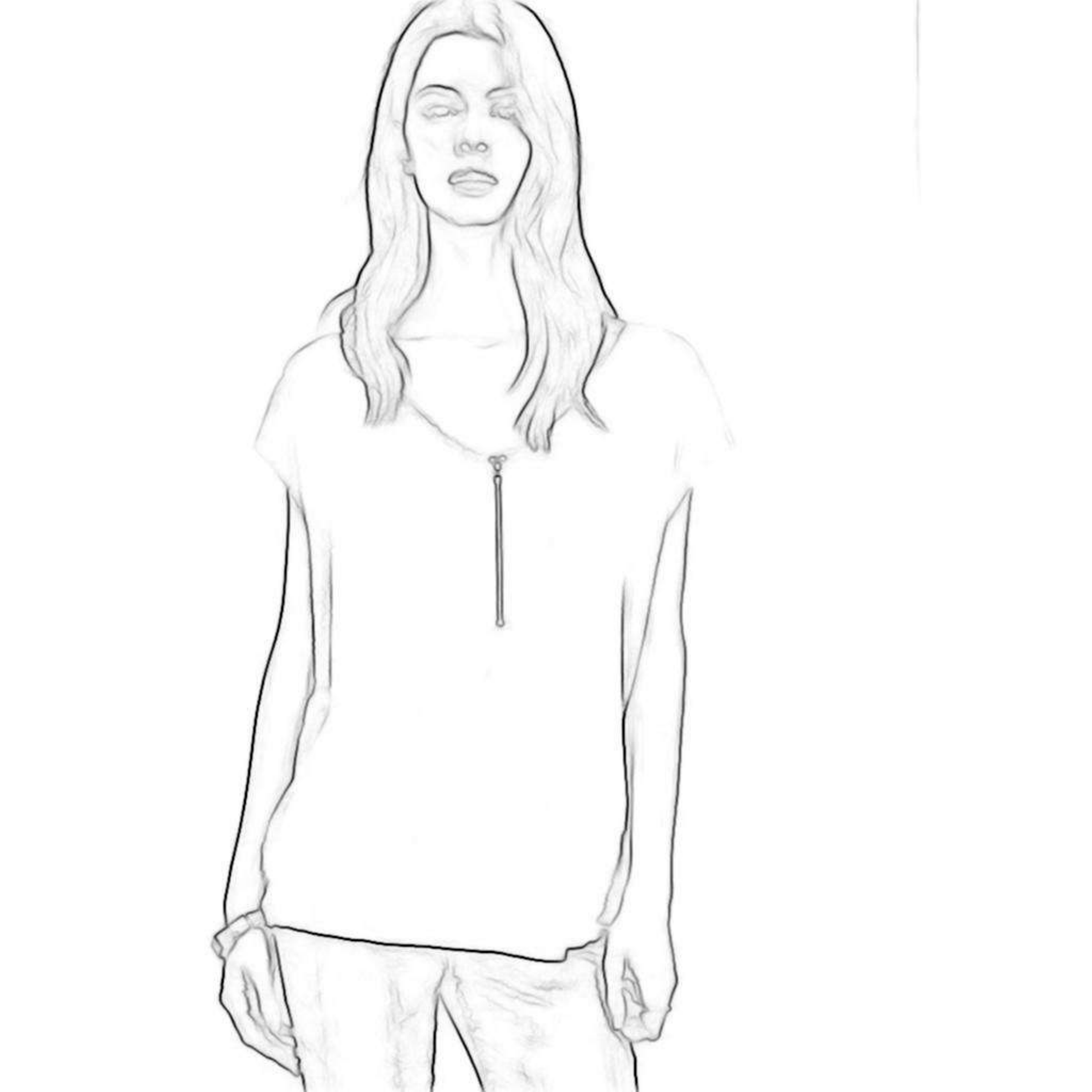}
    \caption{}
    \end{subfigure}
    \begin{subfigure}{.115\textwidth}
    \includegraphics[width=0.8in]{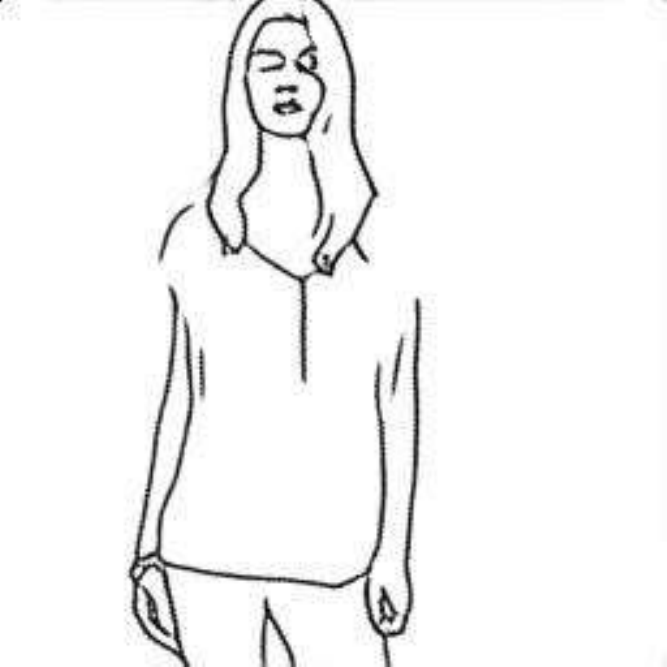}
    \caption{}
    \end{subfigure}
    \begin{subfigure}{.115\textwidth}
    \includegraphics[width=0.8in]{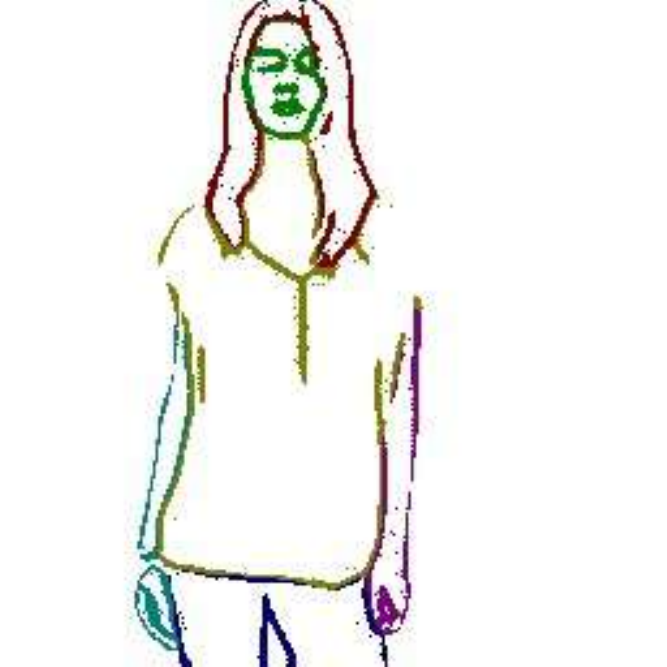}
    \caption{}
    \end{subfigure}
  \caption{The process of building our training and validation sets of sketches. (a): Input human image. (b): Edge extraction of (a) by Im2Pencil \cite{li2019im2pencil}. (c): Sketch simplification of (b) by the method of Simo-Serra et al. \shortcite{simo2018mastering}. (d): Part segmentation of (c) by PGN \cite{gong2018instance}.}
    \label{fig:datset}
\end{figure}

Training the global synthesis network $G$ needs a dataset of paired images and sketches. Similar to previous methods \cite{isola2017image,sangkloy2017scribbler,chen2020deepfacedrawing}, we extract edge maps from human images of $256\times256$ resolution in DeepFashion \cite{liu2016deepfashion} to build our synthetic sketch dataset. At first, we filter the DeepFashion dataset to remove images of the lower body. Then we apply the edge detection method proposed by Im2Pencil \cite{li2019im2pencil} to get an edge map for each human image (Figure \ref{fig:datset} from (a) to (b)). By employing the sketch simplification method proposed by Simo-Serra et al. \shortcite{simo2018mastering}, we %can 
clean noise curves in the extracted edge maps (Figure \ref{fig:datset} (c)) so they resemble hand-drawn sketches more. This results in 
a new large-scale sketch dataset of human images with paired data. This dataset contains $37,844$ pairs in total. We randomly select $2,000$ pairs as the validation set and the remaining $35,844$ pairs as the training set.

Our models also require human parsing maps and pose heatmaps for training. We utilize PGN \cite{gong2018instance} to predict a 
parsing map for each human image in our dataset. To simplify the problem, we merge several labels in the parsing maps, resulting in $C=8$ types of body parts altogether.
% , namely hair, face, top-clothes, bottom-clothes, left and right arms, left and right legs. 
The merged parsing maps are regarded as the ground-truth. These maps also allow us to segment the paired edge maps to obtain semantically segmented edge maps (Figure \ref{fig:datset} (d)). % \hbc{why not remove the noise introduced by part segmentation by using the input sketch as a mask?} % it to segment the sketch, obtaining the sketch segmentation map. 
To prepare the data for training the transformer network, we first employ OpenPose \cite{cao2018openpose} to predict the 2D pose keypoints from the human images, and then % Then we 
generate pose heatmaps from the keypoints based on the Gaussian distribution to better capture spatial features. % to train our models. Figure \ref{fig:datset} illustrates the process of building our training and validation sets.

To evaluate the usefulness of our method in practice, we have collected freehand sketches from 12 users (6 males, 6 females).
%The users consist of six males and six females. 
Four of them have good drawing skills, while the others are less proficient.
% (including 6 males and 6 females of different professional abilities in drawing). \hbc{can you make it clearer, e.g., how many are good at drawings?}
The users were asked to imitate a given human image or just draw an imagined human. They were instructed to draw %drew 
a segmented sketch part by part, 
% \hbc{how the segmentation was annotated? they were instructed to draw part by part or they draw sketches first and then manually segment them?} 
taking around one minute to complete one sketch on average. We have collected 308 % \hbc{better to highlight all the placeholders. otherwise you might forget to replace them in the last minutes.} 
% \hbc{have you tried to generate images for all of them?}
hand-drawn sketches of human images in total to construct our test set. We plan to release our dataset of paired human images and synthetic edge maps as well as hand-drawn sketches publicly for future research.

\subsection{Implementation details}
\label{sec:implementation}

In the geometry refinement module. We share the left and right arms/legs with the same auto-encoders by leveraging the human body symmetry, so there are in total $6$ part auto-encoders. Each part encoder $E^c$ contains five downsampling convolutional layers, with each downsampling layer followed by a residual block. A fully-connected layer is appended in the end to encode the features into the latent vector $v^c$ of $512$ dimensions. Similarly, the part decoders $D_S^c$ and $D_M^c$ each contain five upsampling convolutional layers and five residual blocks in total. The final convolutional layers in $D_S^c$ and $D_M^c$ reconstruct the part sketch $S^c$ and the part mask $M^c$, respectively. %accordingly.
To train the structure refinement module, we preprocess the training set by applying random affine transformations, which are composed of translation, rotation, resizing, and shearing transformations. The spatial transformer network $T_j$ in each step
% \hbc{Do you have one transformer network for each component? If yes, it's better to use $T^c$? In addition, this is not clearly reflected in the pipeline figure (Figure 1).} 
consists of five downsampling convolutional layers, five residual blocks, and the last two fully-connected layers to predict the affine transformation matrices for all body parts.
% To train the global synthesis network $G$, we employ two-scale discriminators \cite{wang2018high} based on the Patch-GAN \cite{li2016precomputed} architecture for the adversarial loss. The perceptual loss measures the distance of layers in the pretrained VGG-19 \cite{simonyan2014very} network, and the feature matching loss \cite{wang2018high} measures the distance of layers in the discriminators.

We use the Adam \cite{kingma2014adam} solver to train all the networks. We set the learning rate to $0.0002$ initially and linearly decay it to $0$ after half iterations. For each part auto-encoder, we first train the encoder $E^c$ and the sketch decoder $D_S^c$ for $100$ epochs and then train the mask decoder $D_M^c$ for $50$ epochs. We train the pose estimation network $P$ and the cascaded spatial transformer network $T$ both for $50$ epochs. We set the batch size to $16$ for the above networks. We train the global synthesis network $G$ for $100$ epochs of batch size $8$ and the face refinement network $F$ for $10$ epochs of batch size $10$. We conduct the experiments by using an Intel(R) Core(TM) i7-4770 CPU @ 3.40GHz with 4 cores and NVidia GTX 1080 Ti GPUs. Please refer to the supplementary materials for more training and architecture details.

\subsection{Comparison with state-of-the-art methods}
\label{sec:comparison}

\begin{figure*}
\centering
\includegraphics[width=1.0\textwidth]{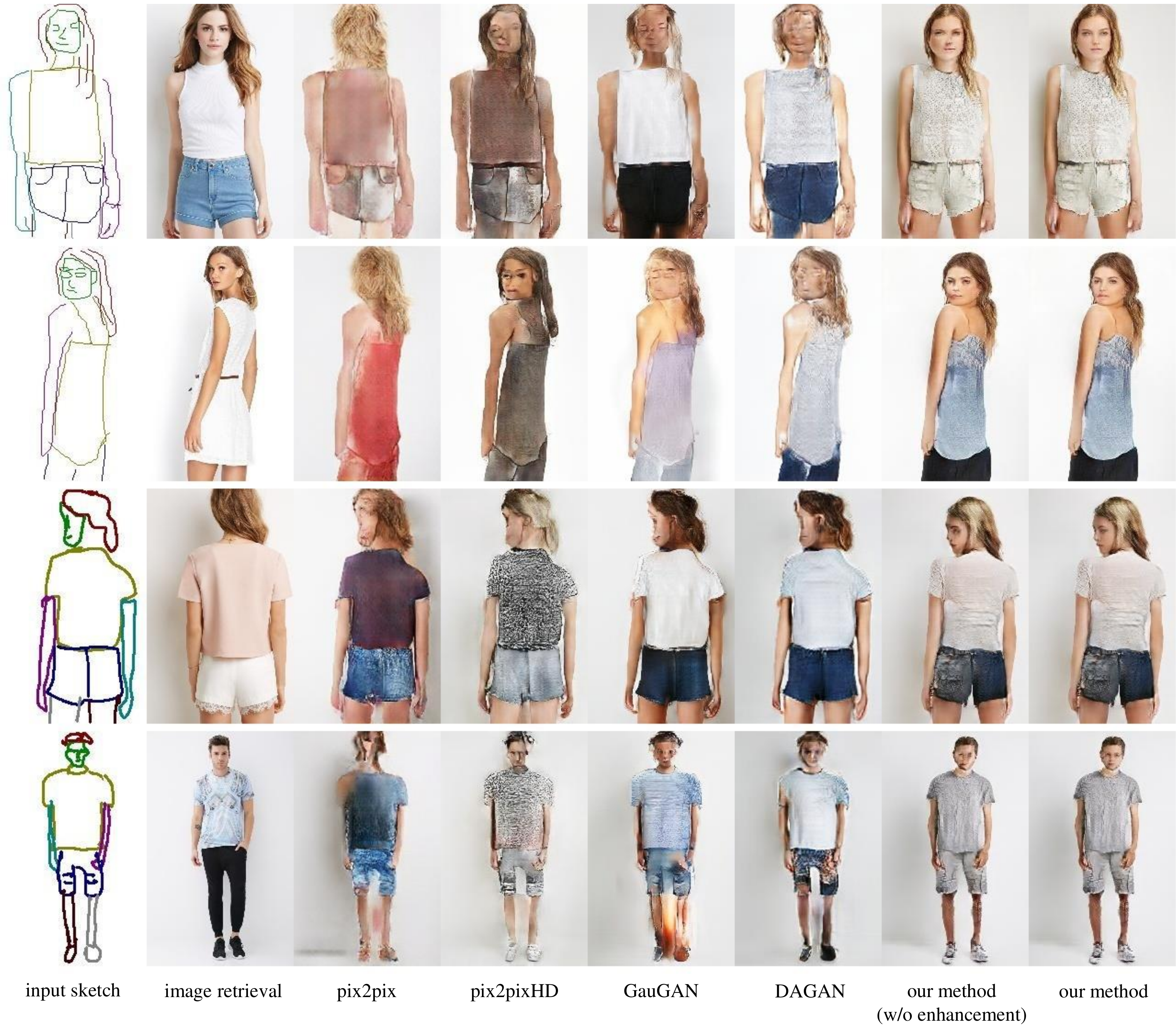}
\caption{Comparison results with a
sketch-based image retrieval method and four state-of-the-art sketch-based image synthesis methods \cite{isola2017image,wang2018high,park2019semantic,tang2020dual}. Our method can produce visually more pleasing results even if the face enhancement module is removed.}
\label{fig:comparsion}
\end{figure*}

To demonstrate the effectiveness of our method for synthesizing realistic human images from freehand sketches, we compare our method with four state-of-the-art sketch-based image synthesis methods, including pix2pix \cite{isola2017image}, pix2pixHD \cite{wang2018high}, GauGAN \cite{park2019semantic} and DAGAN \cite{tang2020dual}. For a fair comparison, we train all the four models on our training set for the same epochs as our method. Please note that we %just 
employ the first-stage generator of pix2pixHD \cite{wang2018high}, since the image resolution of our dataset is limited to %in 
$256 \times 256$. We also compare our method with a sketch-based image retrieval approach. To achieve this, we train an auto-encoder for an entire edge map and collect all latent vectors in the training set. Given an input sketch, we encode it into a vector and retrieve the nearest neighbor from the training set. We regard the human image corresponding to the nearest vector as the retrieval result. To eliminate the influence of facial areas, we remove the face enhancement module in our method for comparison.

Figure \ref{fig:comparsion} shows several representative results of our method and the other five approaches on our test sketches. Compared to the four state-of-the-art sketch-to-image synthesis techniques, our method performs much better with visually more pleasing results. Even when the face enhancement module is removed, our method still produces more realistic texture details and more reasonable body structures, owing to the geometry and structure refinement guided by the semantic parsing maps. Compared to the sketch-based image retrieval approach, our method can produce brand-new human images which respect user inputs more faithfully. Please refer to the supplementary materials for more comparison results. 
% \hbc{You may refer to this figure in our previous discussion on global image retrieval.}
%meet the requirements of users better.

To further evaluate the results, we have applied FID \cite{heusel2017gans} as a quantitative metric, % to evaluate the results, 
which measures perceptual distances between generated images and real images. Table~\ref{tab:evaluation} shows that our method outperforms the other three sketch-to-image synthesis methods \cite{isola2017image,wang2018high,park2019semantic}, indicating % with 
more realistic results by our method. However, as claimed by \cite{chen2020deepfacedrawing}, this perceptual metric might not measure the quality of results correctly, since it does not take the geometry and structure of the human body into consideration. Therefore, we also conducted a user study to compare our method with the three sketch-to-image synthesis techniques \cite{isola2017image,wang2018high,park2019semantic}. We randomly selected 30 sketches from the test set and showed each sketch along with the four results by the compared methods in a random order to users, who were %are
asked to pick the most realistic results. There were 17 participants in total, resulting in 510 votes. Our method received significantly more votes than the other methods, % outperformed the alternative ones receiving most of the votes, 
as shown in Table~\ref{tab:evaluation}. The participants were %are 
also asked to give a score of faithfulness for each result by GauGAN \cite{park2019semantic} (we select it as the representative one of the sketch-to-image synthesis methods), the sketch-based image retrieval method, and our method. The scores ranged %Scores range 
from 1 to 10, the higher the better. Table~\ref{tab:evaluation} shows that the results of our method conform with %are more conform to 
input sketches better than the image retrieval method and are comparable to GauGAN \cite{park2019semantic}. For a fair comparison, we also removed the face enhancement module in our method to produce the results used in %of all results during 
the user study.

% We have tried to use quantitative metrics to evaluate our method, such as FID \cite{heusel2017gans} and IS \cite{salimans2016improved}. However, we find that they cannot measure the quality of results correctly, since they do not take the geometry and structure of the human body into consideration. Instead, we have conducted a user study to compare our method with the three sketch-to-image synthesis techniques \cite{isola2017image,wang2018high,park2019semantic}. We randomly select 30 sketches from the test set and show each sketch along with the four results in a random order to users, who are asked to pick their favourite results. There were 32 participants in total, resulting in 960 votes. Our method outperformed the alternative ones receiving 45.93\% of the votes, while pix2pix\cite{isola2017image}, pix2pixHD\cite{wang2018high}, and GauGAN\cite{park2019semantic} received 9.27\%, 20.94\%, and 23.85\% of votes,  respectively.

\begin{table}[tp]
\caption{Quantitative evaluation of our method, three sketch-based image synthesis methods \cite{isola2017image,wang2018high,park2019semantic}, and an image retrieval method. We have used FID \cite{heusel2017gans} as a quantitative metric and conducted a user study to evaluate the realism and faithfulness of the results. The arrow after each metric identifies the improvement direction.}
\centering
\begin{threeparttable}
\begin{tabular}{cccc}
\toprule
 & FID ($\downarrow$)  & Realism ($\uparrow$) & Faithfulness ($\uparrow$) \cr
\midrule
Pix2pix & 71.12 & 7.65\% & / \cr
Pix2pixHD & 70.87 & 21.37\% & / \cr
GauGAN & 51.92 & 24.71\% & \textbf{6.21} \cr
Image retrieval & / & / & 5.18 \cr
Our method & \textbf{50.36} & \textbf{46.27\%} & 6.15 \cr
\bottomrule
\end{tabular}
\end{threeparttable}
\label{tab:evaluation}
\end{table}

\subsection{Ablation study}
\label{sec:ablation}

\begin{figure}
\centering
\includegraphics[width=0.48\textwidth]{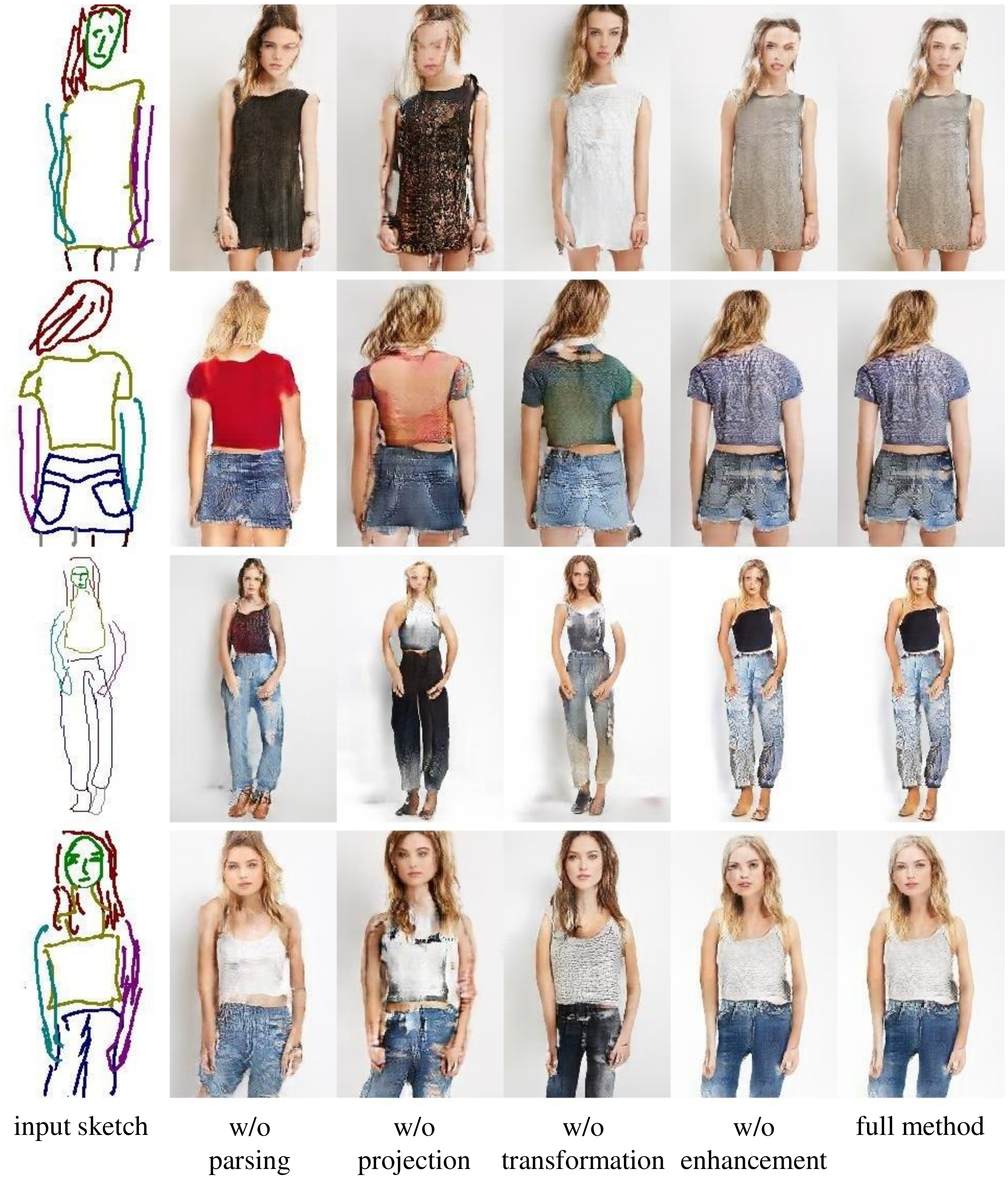}
\caption{Comparison results in the ablation study. We remove the parsing map guidance, the projection of latent vectors, the spatial transformation, and the face enhancement in our method, respectively.}
\label{fig:ablation}
\end{figure}

We have conducted an ablation study to demonstrate the contributions of the different components of our method. Each time, we remove the parsing map guidance, the projection of latent vectors, the spatial transformation, and the face enhancement, respectively, while keeping the other components unchanged. As shown in Figure~\ref{fig:ablation}, without the guidance of the human parsing map, our method cannot produce locally consistent results in the same semantic regions (e.g., legs in the second and third rows). Without the projection component, our method cannot refine the geometry of local details, resulting in obvious artifacts. Without the spatial transformation component, our method will produce results with incorrect connection relationships of joints (e.g., shoulders in the second and third rows) or unreasonable body proportions (e.g., the first and fourth rows). Without the face enhancement, our method may not generate realistic facial details. %landmarks.

\subsection{Multi-modal synthesis}
\label{sec:multi-modal}

\begin{figure}
\centering
    \begin{subfigure}{.48\textwidth}
    \includegraphics[width=1.\textwidth]{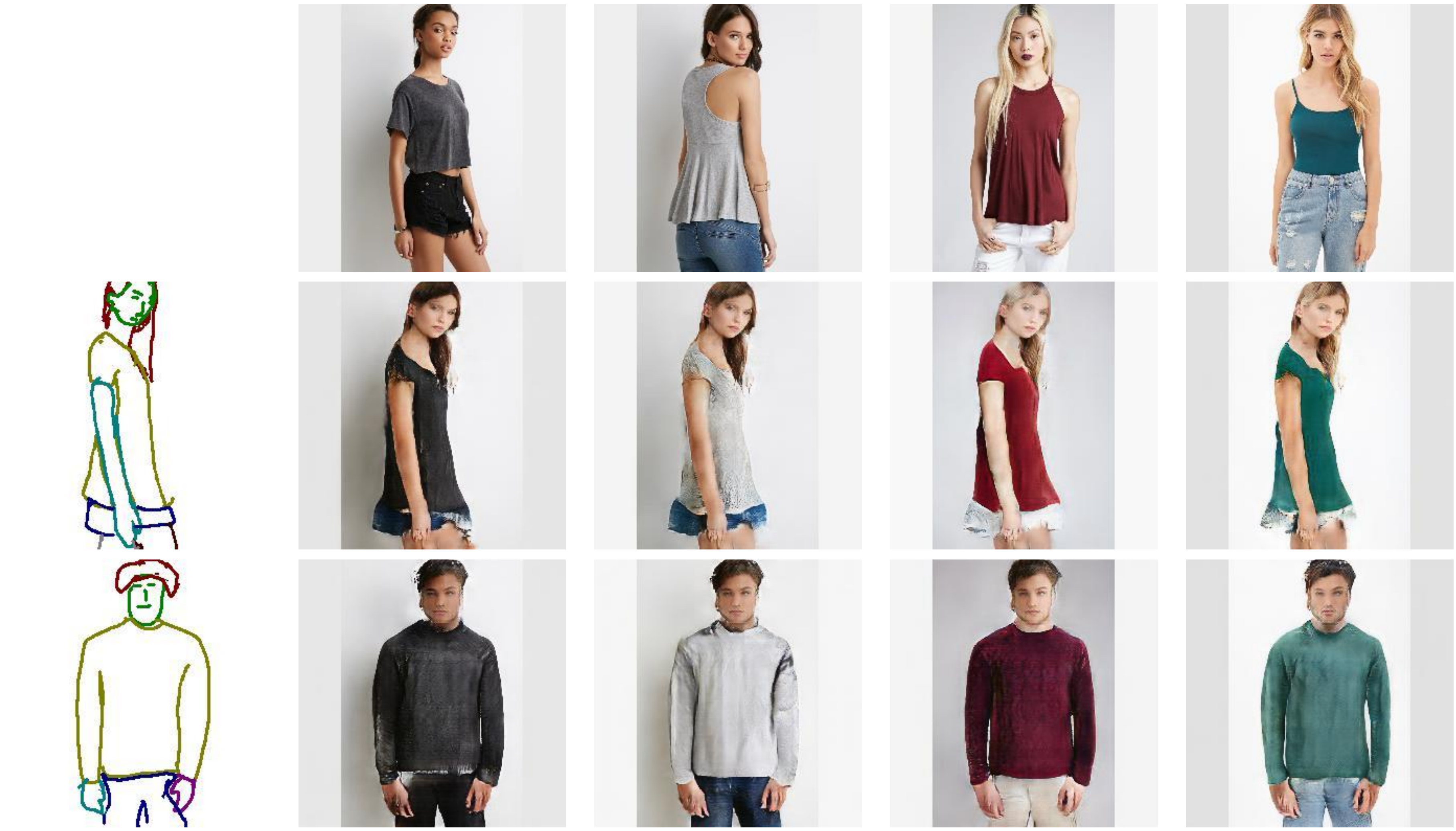}
    \caption{Given reference styles}
    \label{fig:multi-modal-ref}
    \end{subfigure}
    \\
    \begin{subfigure}{.48\textwidth}
    \includegraphics[width=1.\textwidth]{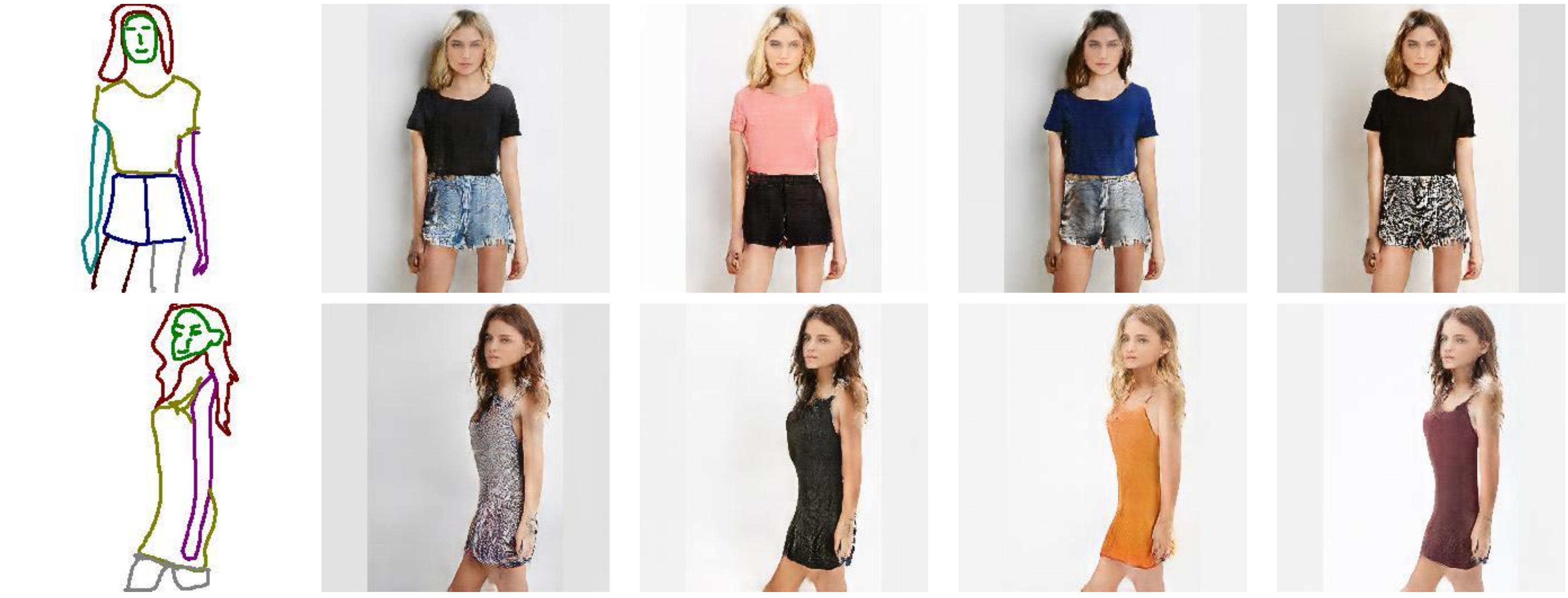}
    \caption{Given random styles}
    \label{fig:multi-modal-rand}
    \end{subfigure}
\caption{For a given input sketch, our method can generate multiple results with texture styles similar to the reference images (a) or random styles (b).}
\label{fig:multi-modal}
\end{figure}

Similar to previous image-to-image translation methods \cite{wang2018high,park2019semantic}, our method can be easily extended % is also able 
to generate multi-modal human images from the same input sketches. To achieve this, %More specifically, 
we append an image encoder ahead of the global synthesis network $G$ and train both of them together with an extra KL-divergence loss \cite{kingma2013auto}. The feature vector encoded by the image encoder can control the texture style of a generated image. Therefore, % given random feature vectors, our method can produce multi-style human images (\wx{Figure \ref{}}). In a similar manner, 
given the feature vectors encoded by reference human images, our method can produce human images with texture styles similar to the reference images (Figure \ref{fig:multi-modal-ref}). Besides, given random feature vectors, our method can also produce diverse human images with different texture styles (Figure \ref{fig:multi-modal-rand}).

%% file: conclusion.tex
% !TEX root = ..\cvpr.tex

\section{Conclusion and Future Work} %s and Limitations

\begin{figure}
\centering
\includegraphics[width=0.45\textwidth]{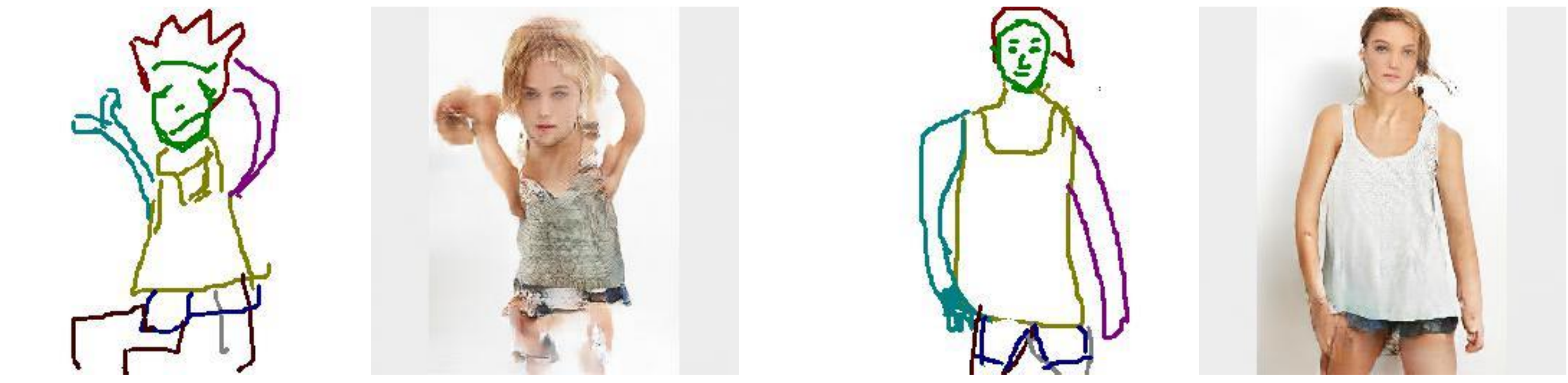}
\caption{
Less successful cases of our method. Left: our method trained on adult images cannot handle a sketched child well. Right: our method trained on images with mixed genders might fail to respect the gender of an input sketch.
}
\label{fig:limitations}
\end{figure}

We have proposed a \textit{projection-transformation-reconstruction} approach %a novel approach 
for generating realistic human images from hand-drawn sketches. Our method consists of three modules, including a geometry refinement 
%sketch conversion 
module, a structure refinement %topology reconstruction 
module, and an image generation module. The geometry refinement %sketch conversion 
module plays an important role in converting roughly drawn sketches into semantic sketch maps, which are locally similar to the edge maps of real human images. This successfully bridges the gap between realistic images and freehand sketches. The structure refinement module locally adjusts spatial connections between body parts and their relative proportions to get a globally more consistent structure. 
%converts the hand-drawn sketch into the part-based sketch map and the parsing map with the geometric constraints. The topology reconstruction module transforms the part maps to recover a reasonable human body structure, by the guidance of the joint relationships. 
The image generation module produces visually pleasing human images with fine facial details. Comparison experiments 
have shown that our approach outperforms three state-of-the-art %previous
sketch-to-image synthesis methods, which %that 
cannot address freehand sketches well. %the global structure and local semantic details of the human body. 
%Ablation study have demonstrated the effectiveness of several modules in our pipeline. Our approach can also produce multi-modal synthesis images with multiple styles. 

Still, the geometry and structure refinement modules %sketch conversion module and the topology reconstruction module 
are restricted to the data distribution in the training set. Therefore, our method cannot produce human images which are very different from the images in DeepFashion~\cite{liu2016deepfashion}. For example, as shown in Figure~\ref{fig:limitations} (Left), our method generates an unsatisfying result for a hand-drawn sketch of a child. The structure refinement %topology reconstruction
module is also limited to recover the human body structure of an adult only since there are only adult models in DeepFashion \cite{liu2016deepfashion}. % \hbc{Can you discuss the influence of different viewpoints and give some results? Not necessarily as a limitation if the results are good. In addition, another limitation is to synthesize loosely dressed human images?} 
As we do not divide the latent vectors of different genders for retrieval, our method is sometimes confused with the gender, as shown in Figure~\ref{fig:limitations} (Right).
We will collect more types of human images to improve the generalization ability of our method in the future work. 
It will also be interesting
to introduce colorful strokes to control the texture styles more exactly.

% \hbc{Can you clean up the references a bit? For example, it's better to refer to the names of journals and conferences in a more consistent format. }